%% file: main.tex
\documentclass[sigconf]{acmart}

\usepackage{booktabs} 
\usepackage{algorithmic}
\usepackage{todonotes}
\usepackage{graphicx}
\usepackage{multirow}
\usepackage{subcaption}
\usepackage{balance}

\setcopyright{rightsretained}

\acmDOI{10.475/123_4}

\acmISBN{123-4567-24-567/08/06}

\acmConference[KDD'19]{25TH ACM SIGKDD CONFERENCE ON KNOWLEDGE DISCOVERY AND DATA MINING}{August 2019}{Anchorage, Alaska USA}
\acmConference[MLG'19]{Workshop on Mining and Learning with Graphs, co-located with KDD '19}{August 2019}{Anchorage, Alaska USA}
\acmYear{2019}
\copyrightyear{2019}

\acmArticle{4}
\acmPrice{15.00}

\begin{document}
\title{Graph Embeddings at Scale}

\author{C. Bayan Bruss}
\affiliation{%
  \institution{Capital One}
  \streetaddress{1680 Capital One Drive}
  \city{McLean}
  \state{Virginia}
  \postcode{22102}
}
\email{bayan.bruss@capitalone.com}

\author{Anish Khazane}
\affiliation{%
  \institution{Capital One}
  \streetaddress{1680 Capital One Drive}
  \city{McLean}
  \state{Virginia}
  \postcode{22102}
}
\email{anish.khazane@capitalone.com}

\author{Jonathan Rider}
\affiliation{%
  \institution{Capital One}
  \streetaddress{1680 Capital One Drive}
  \city{McLean}
  \state{Virginia}
  \postcode{22102}
}
\email{jonathan.rider@capitalone.com}

\author{Richard Serpe}
\affiliation{%
  \institution{Capital One}
  \streetaddress{1680 Capital One Drive}
  \city{McLean}
  \state{Virginia}
  \postcode{22102}
}
\email{richard.serpe@capitalone.com}

\author{Saurabh Nagrecha}
\affiliation{%
  \institution{Capital One}
  \streetaddress{1680 Capital One Drive}
  \city{McLean}
  \state{Virginia}
  \postcode{22102}
}
\email{Saurabh.Nagrecha@capitalone.com }

\author{Keegan E. Hines}
\affiliation{%
  \institution{Capital One}
  \streetaddress{1680 Capital One Drive}
  \city{McLean}
  \state{Virginia}
  \postcode{22102}
}
\email{keegan.hines@capitalone.com}

\renewcommand{\shortauthors}{C. B. Bruss et al.}

\begin{abstract}
   \input{section-abstract.tex}
\end{abstract}

%
%



\maketitle

\input{section-intro.tex}
\input{section-relatedwork.tex}
\input{section-proposed.tex}
\input{section-results.tex}
\input{section-conclusion.tex}

\bibliographystyle{ACM-Reference-Format}
\bibliography{bibliography-biblatex}
\balance

\end{document}

%% file: section-abstract.tex
Graph embedding is a popular algorithmic approach for creating vector representations for individual vertices in networks. Training these algorithms at scale is important for creating embeddings that can be used for classification, ranking, recommendation and other common applications in industry. While industrial systems exist for training graph embeddings on large datasets, many of these distributed architectures are forced to partition copious amounts of data and model logic across many worker nodes. In this paper, we propose a distributed infrastructure that completely avoids graph partitioning, dynamically creates size constrained computational graphs across worker nodes, and uses highly efficient indexing operations for updating embeddings that allow the system to function at scale. We show that our system can scale an existing embeddings algorithm - skip-gram - to train on the open-source Friendster network (68 million vertices) and on an internal heterogeneous graph (50 million vertices). We measure the performance of our system on two key quantitative metrics: link-prediction accuracy and rate of convergence. We conclude this work by analyzing how a greater number of worker nodes actually improves our system's performance on the aforementioned metrics and discuss our next steps for rigorously evaluating the embedding vectors produced by our system.

%% file: section-intro.tex
\section{Introduction}
Graphs are expressive data structures that have been used to represent data in many different real-world applications, including social networks, protein interaction graphs, and communication networks \cite{graphhistory}. All of these representations not only encode critical information about individual vertices but also about a vertex's connections to other entities in its surrounding subgraph. Using machine learning to learn a graph's topology can be useful for predicting unknown relationships between existing and future vertices. However, the size of these graphs typically bottlenecks the effectiveness of running statistical algorithms on them. For example, detecting network motifs on an extremely large graph is computationally expensive and requires an efficient algorithm to scale with the size of an input network \cite{graphhistory}.

A popular approach for learning meaningful representations of vertices on large-scale graphs is by using algorithms to learn graph embeddings \cite{graphhistory}. These embeddings are transformations of graph elements into vector representations, where each vertex's embedding captures its own attributes as well as the local topographic structure in the surrounding subgraph \cite{graphhistory}. This representation works particularly well because vector operations are far simpler and faster than performing matrix multiplications over (say) a sparse adjacency matrix. Common methods for learning graph embeddings include DeepWalk~\cite{DBLP:journals/corr/PerozziAS14}, Node2Vec~\cite{DBLP:journals/corr/GroverL16}, and other methods that use random walks combined with shallow neural networks to produce vector representations. Recently, other attempts have been made to apply variational autoencoders and (VGAE)~\cite{kipf2016variational} and deep neural networks (DNNs) directly to the  whole graph. Both techniques output unique embeddings for every single vertex on a graph, which can then be used for other applications like classification, ranking, or recommendation.

While the aforementioned algorithmic approaches for learning vertex embeddings have improved the quality of these vectors, scaling these methods to work on graphs that have millions of vertices is still an active area of research. In this paper, we propose a distributed architecture that can efficiently use these approaches to learn graph embeddings without being tied to the size or scale of the input network. Our system relies on two key features: entity-based parameter servers and an arbitrary (user-defined) number of worker nodes that communicate with each other to update graph embeddings during the training process. Each worker node retrieves a batch of data from the input graph, requests embeddings for each unique vertex from the correct parameter server, trains on this localized subset, and then updates their respective embeddings by communicating with that parameter server. Furthermore, each worker node dynamically creates a TensorFlow computational graph during each training iteration, and uses a unique indexing approach for mapping local embedding updates to embeddings stored in the centralized parameter server. This technique allows our system to take advantage of highly efficient retrieval and update operations. Finally, this technique allows us to utilize both GPU and CPU resources where most appropriate.

In Section~\ref{sec:relatedwork}, we briefly highlight other infrastructural advancements in current literature that inspired the system proposed in this paper. We then present our own proposed distributed architecture in Section~\ref{sec:proposedapproach}. Following our proposal, we present results from scaling an existing embeddings algorithm - skip-gram - with our distributed architecture on two massive graph datasets in Section~\ref{sec:experiments}: the open-source Friendster graph, a homogeneous social network that contains roughly 68 million users that are linked via 2.5 billion edges, and on an internal heterogeneous graph (IHG) that has two different vertex types with a total of roughly 50 million vertices and 280 million edges. Our results demonstrate that training with our system leads to significant gains in convergence speed while maintaining a high link prediction accuracy, a metric that measures how well an algorithm can infer missing links within an observed network. We conclude this paper in Section~\ref{sec:conclusion} by summarizing our key contributions and identifying future work that expands on the framework laid out by this paper.

%% file: section-relatedwork.tex
\section{Related Work}
\label{sec:relatedwork}

While neural embedding models have effectively created rich mathematical representations for words in natural language processing, attempting to model relationships between entities in larger datasets --- such as the links between users in a social network graph --- requires a far more scalable approach. For example, the original Word2Vec~\cite{word2vec} language model creates embeddings for roughly 692,000 different words, whereas training on the Friendster dataset, an open-source social network, requires creating representations for nearly 68 million unique users~\cite{friendster}. Tackling this difference in scale requires a modeling approach that is not tied to the size of the dataset.

In this section, we will examine a few distributed systems proposed in academia and industry that are able to effectively train on datasets with up to hundreds of millions of unique entities. We will primarily analyze how key infrastructural features within these systems allow them to run algorithms at scale.

\subsection{HogWild! Asynchronous Training of Large-Scale Datasets}
One challenge while using stochastic gradient descent (SGD) is parallelizing gradient updates from several points of data during training. This algorithm is inherently serial and training on a large-scale dataset may take an inordinate amount of time if implemented in a non-distributed fashion. Prior to advancements with asynchronous distributed training, most of the best approaches for speeding up serial algorithms required using additional computational power (e.g, GPUs) to iterate a model on larger batches of training data~\cite{recht2011hogwild}. 

\textit{HogWild!} is a distributed methodology for training on large-scale datasets without any synchronization; all threads on a multi-core system can read and write to a global list of model parameters at any time during a training job. Part of \textit{HogWild!}'s success comes from its proof that asynchronous training has no negative impact on the rate of convergence for a large-scale machine learning model, which has been successfully extended to many different applications including stochastic coordinate descent for SVMs and randomized Kaczmarz algorithms for solving systems of linear equations~\cite{de2015taming}. We adopt the \textit{HogWild!} philosophy while training our distributed system on the large-scale datasets in the experiments section of this paper. 

One possible misstep that can arise while using the \textit{HogWild!} philosophy is forcing every worker node to have a replica of the global computational graph during training. Storing a graph embedding model on every worker node that requires access to millions of vector representations would drastically hamper a system's ability to horizontally scale to multiple machines. While we train our distributed system in an asynchronous fashion, we only retrieve and update embeddings for entities within a given \textit{batch} of data, allowing each worker to dynamically execute a computational graph with only the embeddings required for that training iteration. We expand on this approach in greater detail in Section~\ref{subsec:distributedtraining}.

\subsection{Distributed Word2Vec Systems for Training 100 Million-wide Vocabularies}

Constructing a distributed Word2Vec training system that can operate with massive vocabulary sizes - roughly 100 million words - is an active area of research. Recently proposed systems are typically composed of a parameter server that maintains globally shared parameters - e.g embeddings in the case of graphs - as well as several worker nodes that communicate with the server to update these parameters during training~\cite{ordentlich2016network}.

In a large-scale distributed Word2Vec system, the central server stores embeddings for every unique word as rows in a global matrix. This matrix is then columnar partitioned into many different parameter server "shards". Each shard corresponds to a subset of columns for this matrix, or a group of features for all embedding vectors. For example, a vector dimension d = 300 that is columnar partitioned equally into 20 parameter server shards would hold 15 features of an embedding in each shard. Consequently, for $S$ parameter shards, their system is able to scale up vocabulary size by a factor of $S$ relative to training on a single machine. 

Vertically partitioning embedding vectors does improve load balancing and reduces network traffic by distributing the number of worker requests to multiple shards as opposed to a single parameter server. However, the maximum embedding dimension size in current literature is already capped at roughly 500, whereas the number of total unique embeddings for an application will likely be far larger depending on the size of the dataset. Thus, horizontally partitioning the total number of rows in a globally shared embeddings matrix is arguably more effective for training on large-scale datasets that include millions of unique entities. The system proposed in this paper tackles this problem by transforming input graphs into a matrix of random walks and then row-wise partitioning this matrix onto several worker nodes. These nodes only request the embeddings for the entities in their respective batch of data at every training iteration. This approach is highly scalable, and we present an interesting correlation between an increase in the number of worker nodes and link prediction accuracy in Section~\ref{sec:experiments}.

\subsection{Large-scale Recommendation Engines for Millions of Users}

Scaling recommendation engines to make personalized suggestions from a pool of billions of items to millions of users is a common challenge in building robust product recommendation engines in industry. Users and items are typically represented as an object graph, which depending on the online retailer may scale up to billions of vertices and edges~\cite{eksombatchai2018pixie}. Many large-scale recommendation engines take advantage of this graphical structure by computing random walks to select the most relevant items for a particular user from a sequence of transactions.

These large-scale recommendation engines have several key features that allows them to efficiently run random walks on the scale of billions of vertices. Firstly, the billion-wide object graph is represented as an array, which allows a model to use highly efficient array operations to find vertices on the graph. This is particularly important for running random walks on billions of vertices as finding the next neighbor in a hop becomes a constant-time lookup operation. Secondly, these systems typically use a distributed cluster in order to parallelize random walks across many different machines. Each machine also stores a copy of the pruned billion-wide object graph, which allows each one of the cluster's workers to independently run random walks without crossing over into other machines. Both of these features bring huge performance benefits for these systems, and our own approach uses a similar indexing technique along with a distributed cluster to run graph embeddings algorithms on massive datasets.

While large-scale recommendation engines are able to scale their systems to operate on billions of vertices, many engines do not produce embeddings for each object in their billion-wide object graphs and still use costly graph partitioning techniques to allocate data to different worker nodes. Therefore, these systems must re-run their random walk approach at every user request in order to find new recommendations. Our goal is to build a similar system that can produce embeddings for every arbitrary object in an input graph and avoid partitioning an input graph during training. These embeddings could then be used for more than just recommendation, but also for classification, anomaly detection, and many other applications.

\subsection{Billion-scale Graph Embeddings System for E-Commerce Recommendation}
Online marketplaces are another type of retailer that frequently interacts with a large number of users and items. One particular architecture, coined the Base Embeddings System (BES), learns embeddings for nearly two billion items on an e-commerce platform~\cite{wang2018billion}. The BES architecture specifically runs DeepWalk on this massive graph to learn the latent representations of each item, and can run at scale due to a few important architectural features.

Firstly, the billion-wide items graph is split into several subgraphs of roughly 50 million vertices per group, and then sent to BES's Open Data Processing Service (ODPS) distributed platform for further processing. The ODPS framework contains thousands of servers distributing as clusters in different regions. Therefore, the BES architecture is able train on each subgraph in a separate cluster, allowing the system to reduce the overall training time by parallelizing DeepWalk across multiple machines. Furthermore, ODPS has many different characteristics that makes it particularly apt for training graph embeddings at scale. It's extremely fault tolerant, which allows the system to re-create workers in case of failures, and the framework is able to efficiently store massive amounts of data by partitioning it across multiple machines. Additionally, running a job with more computational power simply requires adding more machines.

While the BES's highly distributed infrastructure is central to running embedding algorithms at scale, its architecture arguably falls short in two areas. Firstly, the system uses an adjacency matrix to represent links between vertices. Navigating a list during a random walk will scale linearly with the number of vertices on that walk, which is further worsened by running numerous random walks for every vertex as specified in the DeepWalk algorithm. While our system also runs random walks as an initial step for learning embeddings, we use an array indexing approach for efficiently retrieving the location of a vertex and its neighbors along any walk. 

Secondly, BES partitions an input graph into several subgraphs at random - without giving much importance to the specific edges that are pruned - and individually trains on each group without centralizing updates to a global parameter server. Not aggregating embedding updates from multiple clusters arguably biases vertices' vector representations to only encode topographic information about edge relationships in their immediate vicinity. Thus, this lack of parameter centralization can make it more difficult for predicting relationships between vertices that are in separate subgraphs. We avoid this problem by using alternative approach to partitioning input graphs, which we describe in Section~\ref{subsec:embeddingalgorithm}.

%% file: section-proposed.tex
\section{Proposed Approach}
\label{sec:proposedapproach}

Embedding a graph in dense vector space is often accomplished by joining two components: a means to encode the local neighborhood of a node (e.g. random walk algorithm) and a means for dimensionality reduction (e.g. a neural embedding model)~\cite{DBLP:journals/corr/GroverL16, DBLP:journals/corr/PerozziAS14}. The random walk algorithm can take a variety of forms and constraints but generally speaking it is used to define a training set for the neural embedding model. This training set comprises a series of examples of vertices that should be similar in the embedding space and often includes examples of vertices that should be dissimilar. From an infrastructure perspective we take a decoupled approach, first generating the sequence as a training dataset followed by learning on the sequences themselves. This allows us to utilize the significant performance enhancements that have already been established for random walk algorithmic implementations. At the same time, it allows us to focus on solving the various constraints of scaling neural embedding models to billions of vertices.

\subsection{Embedding Algorithm}
\label{subsec:embeddingalgorithm}
The standard neural embedding model used is the skip-gram Word2Vec model. The Word2Vec model words are embedded in dense vector space by sliding a window over sequences of words in a corpus. For each word the task then becomes using the vector representation of each word to predict the words immediately before and after it in the window~\cite{NIPS2013_5021}. While the exact formulation for this task requires an expensive computation of a softmax over all possible vocabulary words, two approximations have gained prominence in practice: hierarchical softmax and noise contrastive estimation (NCE) \cite{mnih2013learning, DBLP:journals/corr/cs-CL-0108006}. 

NCE was first proposed by Gutman et al.~\cite{pmlr-v9-gutmann10a} and then applied to word embeddings by Mnih et al. \cite{mnih2013learning} has gained equal popularity to the hierarchical softmax as an approximation of the softmax. NCE decomposes the skip-gram algorithm into a binary classification task. In this task for an input word $x$, observed context words $Y$ and a set of randomly generated noise samples $N$, the model fits a logistic regression model to the embedding layer such that it can distinguish between the noise samples $N$ and the true context words $Y$ when given $x$.

This model has been further generalized by simplifying the logistic regression task. This is accomplished by directly comparing the vectors of the input word $x$, context word $y$ and noise samples $N$ in a pairwise fashion. The choice of comparison function can vary, but frequently is cosine similarity or another distance metric. When generalized beyond language, the objective of this model is to determine a low dimensional representation of an vertex in a graph such that its distance to observed edges on the graph in the low dimension space is minimized while maximizing its distance from observed non-edges~\cite{DBLP:journals/corr/abs-1709-03856}. This objective has been shown to be factorizing the adjacency matrix~\cite{TACL570}. However, because it explicitly models non-edges in the noise sampling procedure it captures relationships that matrix factorization cannot. 

Finally, for the purpose of the architecture in question, the generalized noise contrastive estimation framework as laid out in Wu et. al. has performance advantages over both hierarchical softmax and traditional noise contrastive estimation. In the case of the former, it is required that during training the entirety of the binary search tree is maintained by the neural network. In a distributed training paradigm this would be computationally efficient but not memory efficient. On the other hand, NCE can be very memory efficient, at each step in training the model only needs the vectors and weights corresponding specifically to the vertices (both positive and negatives) in the training batch. As will be shown further on in this paper, this property allows for significant parallelization. Furthermore, by using the simpler NCE with just the distance metrics, the memory footprint of the model is halved. 

\subsection{Distributed Graph Traversal}
While our implementation allows for a variety of graph traversal approaches, we make a few key observations and use notes. Firstly, some other approaches have proposed graph partitioning~\cite{wang2018billion} as a means of constraining the number of vertices per worker and allowing for parallelization. The benefit of this is that each worker can get a distinct subset of vertices and therefore no centralized parameter store or synchronization. If there are no existing subgraphs then partitioning has to occur by severing edges. Accomplishing a balanced partition that minimizes the edges cut is an NP-hard problem~\cite{Tsourakakis:2014:FSG:2556195.2556213}. While approximations exist, if possible we find it better to use other mechanisms for parallelizing the training of this model that don't involve partitioning the graph. For example, approaches such as those described in Eksombatchai, et al.~\cite{eksombatchai2018pixie} allow for the graph algorithmic portion of the embedding model to be performed in-memory on a single vertex even on very large (3 billion vertex) graphs.

Secondly, because we use noise contrastive estimation, we constrain our graph walk algorithm to have a small context window, only modeling first order relationships between vertices. The motivation here is to guarantee true noise samples. If ``context'' is defined as a shared edge, for any given vertex any non-edge becomes a valid ``noise'' sample. When using NCE, if context is defined, as it is in DeepWalk, as five steps away on the graph then it becomes difficult to guarantee that a randomly generated noise vertex will not appear within five steps of that given vertex. Depending on the graph at a certain context size, everything becomes inter-related and the idea of noise sampling becomes irrelevant.

The number of training pairs generated by graph traversal as a function of walk length ($l$), context window ($c$) and walks per vertex ($w$) is:

\begin{align*}
  \text{pairs} = w*\sum_{j=2}^{c} (l-j+1)  
\end{align*}

If we constrain $c=2$ then this reduces to $\text{pairs} = w*(l-1)$. With this constraint in place, any reduction in walk length can then be offset by taking more walks per vertex.



\subsection{Distributed training}
\label{subsec:distributedtraining}

For a typical large graph, the number of training pairs generated by a graph walk algorithm may far exceed what can processed by a neural embedding model on a single machine. At the same time it can be observed that a limited number of vertices appear on any given subset of training data. Thus, parallelization can be achieved by splitting the training set across a number of workers that execute in parallel asynchronously and operate relatively independent of each other. Each worker fetches a batch of data, trains and updates the embeddings of only those vertices which appear within its assigned batch. Distributed training relies on a centralized parameter repository that allows learning to be shared across all the workers. In the case of graph embeddings, a parameter server can hold the embeddings table which contains the vectors corresponding to each vertex in the graph. Figure \ref{fig:arch_fig} provides a visual summary of this setup, where Servers 1 through $N$ are global parameter servers that store embedding vectors for $N$ different vertex types in a given heterogeneous graph (vertex type A, vertex type B, etc), while workers 1 through $N$ will communicate with the aforementioned servers for retrieving and updating embedding vectors. Creating $N$ servers for $N$ vertex types decouples the training process for each vertex type entirely; the user can choose to instantiate and train embedding vectors for only a desired combination of vertex types instead of pre-allocating memory space for all possible vertex types. Furthermore, all parameter servers utilize row based index id where the id of a vertex equals the row id in the table. This allows for highly efficient retrievals and updates.
During each training iteration, the workers load a subset (batch) of training data which can contain a variable number of unique vertices. The embeddings of these unique vertices are requested from the parameter servers. Since embedding lookup functions in the Tensorflow require the data to be formatted as an index referencing its location in a Tensor (of maximum size 2GB for GPUs) a mapping is created from a parameter server index to the worker's tensor index. After training takes place on the worker, the embeddings are updated on the parameter servers. In summary, each worker executes the following steps:

\begin{figure}[t]
\centering
\includegraphics[width=8cm]{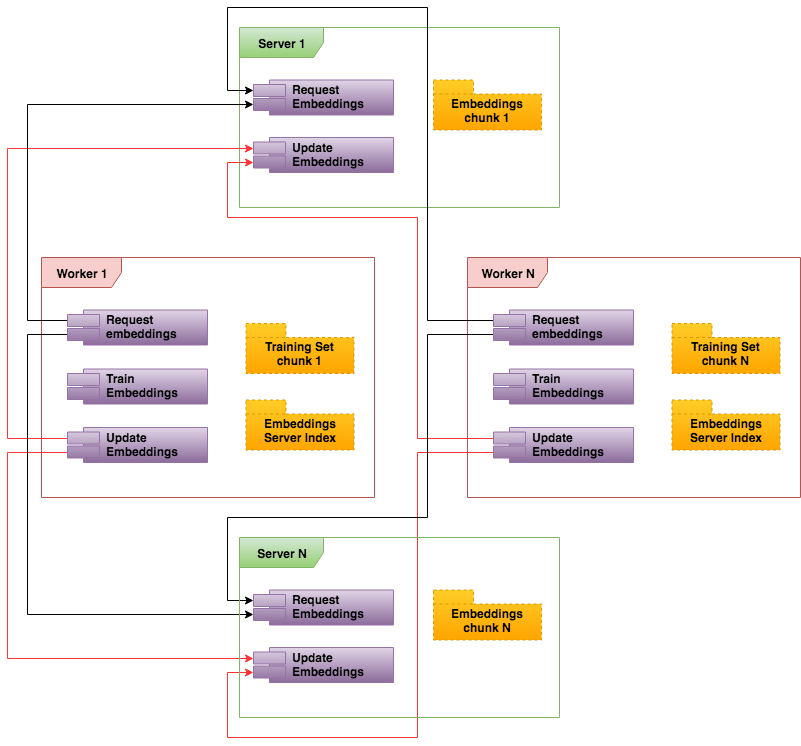}
\caption{Distributed training architecture diagram}
\label{fig:arch_fig}
\end{figure}

\begin{algorithmic}
\STATE Determine size of training subset
\STATE $dataSize\gets nSteps * batchSize$
\FOR{iter in size(trainingData) / dataSize}
    \STATE $subset\gets trainingData / dataSize$
    \STATE $uniquevertices\gets set(subset)$
    \STATE $localIdx\gets 0$
    \FOR{vertex in uniquevertices} 
        \STATE $vertexToLocal[vertex]\gets localIdx$
        \STATE $localTovertex[localIdx]\gets vertex$
        \STATE $localIdx \gets localIdx + 1$
    \ENDFOR
    \STATE Replace ids in subset of data with local indices
    \FOR{row in subset}
        \FOR{element in row}
            \STATE $localElement\gets vertexToLocal[element]$
        \ENDFOR
    \ENDFOR
    \STATE Request existing embeddings from server for unique vertices
    \STATE $initialEmbeddings\gets serverRequest(uniquevertices)$
    \STATE Uses these embeddings as intializations for embeddings tensor in graph
    \STATE $embeddingsTensor\gets initialEmbeddings$
    \FOR{step in nSteps}
        \STATE Train embeddings using batch of subset data
    \ENDFOR
    \STATE Map rows in embeddings tensor back to global values
    \STATE Send trained embeddings back to server to update values
\ENDFOR
\end{algorithmic}

This system for training node embeddings is fully asynchronous without locking as used in Recht, et. al.~\cite{recht2011hogwild}. Much like \textit{HogWild!}, we also assume that any two subsets of data will have a limited number of shared vertices and therefore update conflicts will be limited. If there is a collision then the most recent server request will overwrite the previous update before any other request can use them in training. Over large datasets we expect this to have minimal impact. Another possibility is that while one worker is updating the embedding of a vertex another worker requests that embedding and gets some portion of the older vector and some portion of the one being written. Because the gradients are small between training iterations the difference between the last version the embedding and the updated version should also be limited.

\subsection{Parameter server partitioning}

The embedding dimension ($D$) is a hyperparameter that can be set to any integer value greater than zero. In most practical applications, typical values of $D$ range anywhere from 100 to 300. For very large graphs ($|V| > $ 1 billion), the memory requirements of serving these embeddings can quickly exceed those provided by commodity servers, motivating the use of partitioning strategies. Most commonly, these strategies either partition across the rows (vertices) or the columns (embedding dimensions) of the table. Column wise partitioning, as laid out in~\cite{grbovic2015commerce}, maintains a number of columns for every vertex on each parameter server. This achieves load balancing as every server is queried once by each worker on each training iteration. The disadvantage is that it restricts the maximum number of parameter servers to the number of embedding dimensions. At the same time, it must be ensured that at a minimum, a single column can fit into the memory of every parameter server.

\begin{figure*}[t]
\centering
    \begin{subfigure}[b]{0.45\textwidth}
        \includegraphics[width=\textwidth]{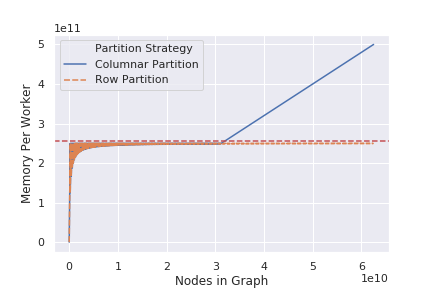}
        \caption{Memory per worker for row wise and column wise partitioning strategies. Red line indicates 256 GB, or standard maximum memory for parameter servers.}
        \label{fig:partition_fig_a}
    \end{subfigure}
    \begin{subfigure}[b]{0.45\textwidth}
        \includegraphics[width=\textwidth]{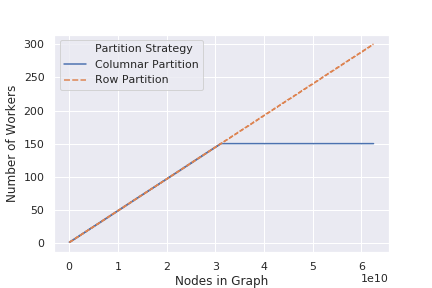}
        \caption{Number of workers required as nodes in graph increases. Column wise partitioning is capped at 150 for illustrative purposes. Embedding dimensions typically don't exceed 300.}
        \label{fig:partition_fig_b}
    \end{subfigure}
\caption{Performance benchmarks on large scale graphs.}
\label{fig:partition_fig}
\end{figure*}

Figure~\ref{fig:partition_fig} shows the limitations of the columnar partition strategy. Namely, that the number of parameter servers cannot exceed the the number of embedding dimensions. In Subfigure~\ref{fig:partition_fig_a} it can be seen that at about 30 billion nodes, no parameter server could hold a single partition. Subfigure~\ref{fig:partition_fig_b} shows that for the row wise partition, the number of workers is able to scale beyond 150 to meet the increasing size of the graph. In practice, it should be acknowledged that very few graphs will reach this size.

For row-wise partitioning, each parameter server holds all the embedding dimensions for a number of vertices. The disadvantage is that parameter servers that hold the embeddings for vertices that appear more often in the training pairs will receive a high number of requests. This can be addressed by utilizing a bin packing approach where the vertices are distributed across the servers in accordance to server capacity and the number of times each vertex appears in the training pairs. Since bin packing is an NP-hard problem, we use the following heuristic solution. We partition embedding vectors by vertex type and use the row-wise partitioning process described earlier in this section. In both approaches, no effort has been made to implement server fault tolerance and the number of servers, graph vertices and training pairs are expected to remain static during training.

%% file: section-results.tex
\section{Experiments and Analysis}
\label{sec:experiments}

In order to evaluate the scalability of our system, we run experiments on two large-scale graph datasets: the Friendster graph and IHG network. Our goal here is to benchmark the efficacy of our proposed technique in two key areas --- 1) the accuracy of the learned embeddings on a typical predictive task (here we use link prediction), and 2) how quickly our distributed architecture achieves a baseline predictive performance on said task.

\subsection{Data and Training} 

\begin{table*}[!htb]
\caption{Dataset Descriptions --- for the Friendster and Internal Heterogeneous Graph (IHG). Note that the IHG has two distinct types of vertices which we outline here.}
\begin{tabular}{lrr}
\hline
 & \multicolumn{1}{c}{\textbf{Friendster}} & \multicolumn{1}{c}{\textbf{IHG}} \\ \hline
\multirow{2}{*}{\# of vertices} & \multirow{2}{*}{68,349,466} & 18,856,021 (vertex type A) \\
 &  & 32,107,404 (vertex type B) \\
\# of edges & 2,586,147,869 & 280,422,628 \\
\# of positive samples per node & 80 & 1 \\
\# of negative samples per node & 400 & 5 \\ \hline
\end{tabular}
\label{table:datasets}
\end{table*}

\begin{table*}[ht]
\caption{Link prediction accuracy results from training on the Friendster and IHG datasets for 700 steps with differing quantities of worker nodes.}
\begin{tabular}{lrrrr}
\hline
\multicolumn{1}{c}{\textbf{Dataset}} & \multicolumn{1}{c}{\textbf{\# of Worker Nodes}} & \multicolumn{1}{c}{\textbf{+ve Accuracy (\%)}} & \multicolumn{1}{c}{\textbf{-ve Accuracy}} & \multicolumn{1}{c}{\textbf{Total Accuracy (\%)}} \\ \hline
\multirow{4}{*}{Friendster} & 7 & 72 & 62 & 67 \\
 & 14 & 74 & 65 & 69.5 \\
 & 21 & 75 & 66 & 70.5 \\
 & 28 & 77 & 67 & 72 \\
IHG & 28 & 95 & 85 & 90 \\ \hline
\end{tabular}
\label{table:accuracies}
\end{table*}

The Friendster graph is a massive open-source dataset that contains 68,349,466 users that are connected to each other with 2,586,147,869 edges. The IHG dataset is an internal heterogeneous graph composed of two vertex types (A and B) that total 50,963,425 total vertices and 280,422,628 edges. We specifically use a skip-gram word2vec implementation to train embeddings.
As our proposed system also uses noise contrastive estimation (NCE) for deriving a loss, positive and negative samples are required to be paired with each node during training. A user's "positive" pair is defined as another user who shares an edge with it, whereas a "negative" pair is defined as another user who is not connected to the original user by an edge. For the Friendster graph, we pair each unique user in the graph with 80 positive and 400 negative pairs. During experimentation, we find that we only need to pair each vertex type in the IHG graph with 1 positive and 5 negative samples to achieve optimal performance.  We summarize the statistics for both datasets in Table \ref{table:datasets}. For evaluation, we use a 90\% / 10\% training-test data split and benchmark our system by running metrics on the held-out test set.

Instead of partitioning an input graph across multiple worker nodes, we pre-compute random walks as specified in Section 3 and construct a data matrix to store these fixed size sequences. We then row-wise partition this matrix across multiple workers during training. 
Furthermore, we train our system on both the Friendster and IHG graphs with up to 28 Nvidia GTX 1080 Ti GPUs depending on the specified number of worker nodes. All experiments are trained until 700 steps (7 epochs). We choose the larger and homogeneous Friendster graph to analyze the trade-offs from training with a variety of different worker nodes in order to focus on the impact of our system's key infrastructural features on its scalability and performance. We also report our results from training on the multi-entity IHG graph in Table \ref{table:accuracies}. We conclude this section by discussing the correlation between link prediction accuracy and worker quantity.

\subsection{Metrics} 

\subsubsection{Link Prediction Accuracy:}

Suppose we have an unweighted and undirected graph $G$ with a vertex set $V$ and an edge set $E$, where an edge $e$ represents a link between two arbitrary nodes $u$ and $v$. The link prediction task is to predict whether there will be an edge \textlangle{}$u,v$\textrangle{} based on the attribute information of both nodes and the topological information in their surrounding subgraph.

If an edge \textlangle{}$u,v$\textrangle{} exists, then we can assume that both $u$ and $v$ are closely related and are thus positive pairs. Otherwise, we assume that they are not related and are negative pairs. Within a held-out set of data, positive accuracy represents the percentage of positive pairs that are correctly identified within the held-out set. Negative accuracy represents the percentage of correctly predicted negative pairs within the same validation data. Total accuracy is simply the average of both the aforementioned accuracies.

During training, we hope that the node attribute and subgraph information for $u$ and $v$ is encoded within their embedding vectors. Thus, given only the embedding vectors for both nodes, our task is to predict whether an edge exists between them or not. We present a breakdown of our system's link prediction accuracy scores in Table \ref{table:accuracies}.

\subsubsection{Rate of Convergence}

We also measure how quickly our distributed architecture takes to cross 70 \% total link prediction accuracy while training on the Friendster graph, with respect to differing amounts of worker nodes. We graphically display how well our system performs on this task in Figure \ref{fig:many_workers_vs_link_fig}.

\subsection{Quantitative Results and Analysis} 

\begin{figure}[ht]
\centering

    \begin{subfigure}[b]{0.45\textwidth}
        \includegraphics[width=\textwidth]{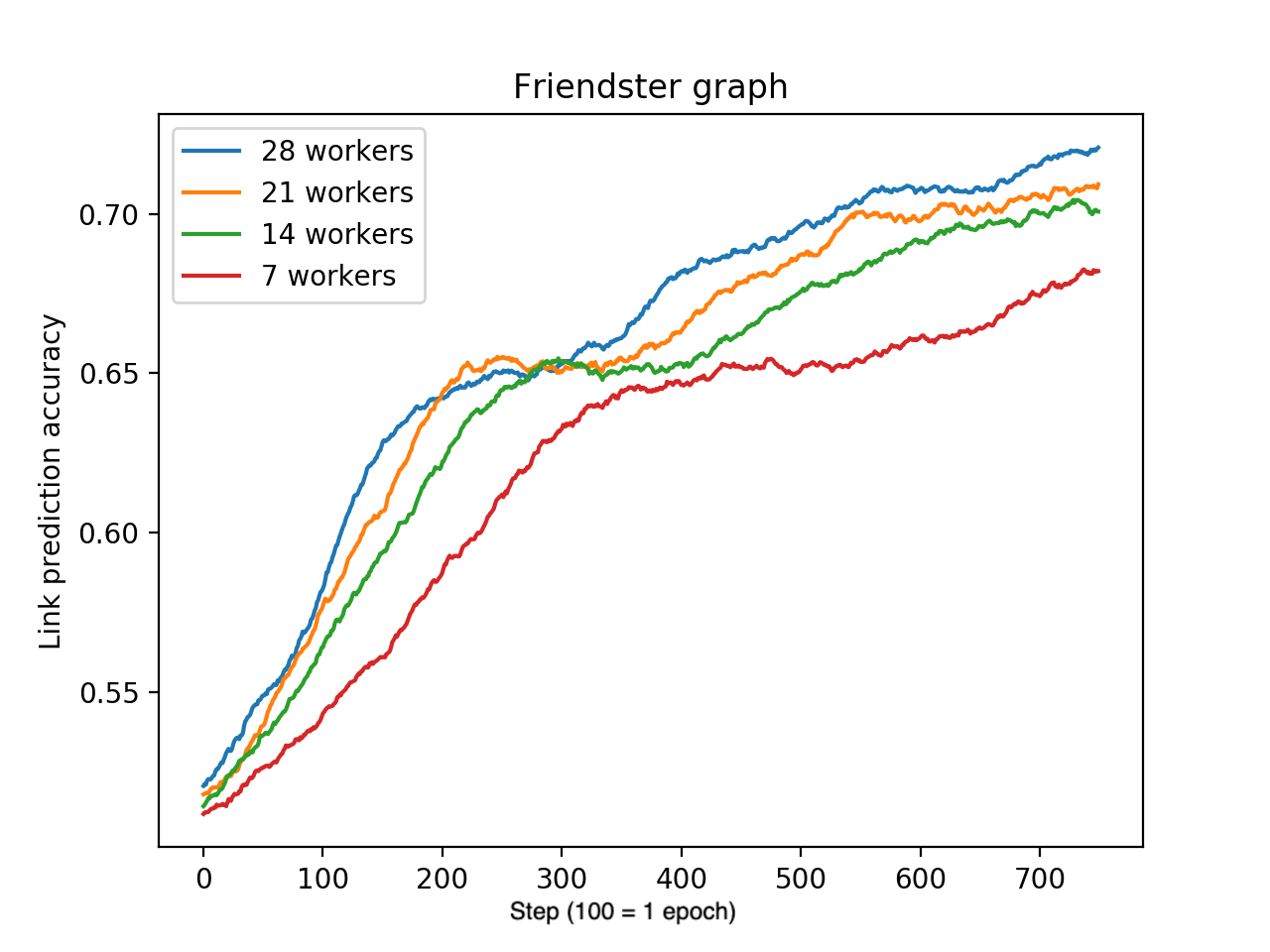}
        \caption{Results from training with a differing number of worker nodes. While all workers converge to the roughly the same link prediction accuracy, more workers allows for faster and more accurate training. }
        \label{fig:many_workers_vs_link_fig}
    \end{subfigure}
    
    \begin{subfigure}[b]{0.45\textwidth}
        \includegraphics[width=\textwidth]{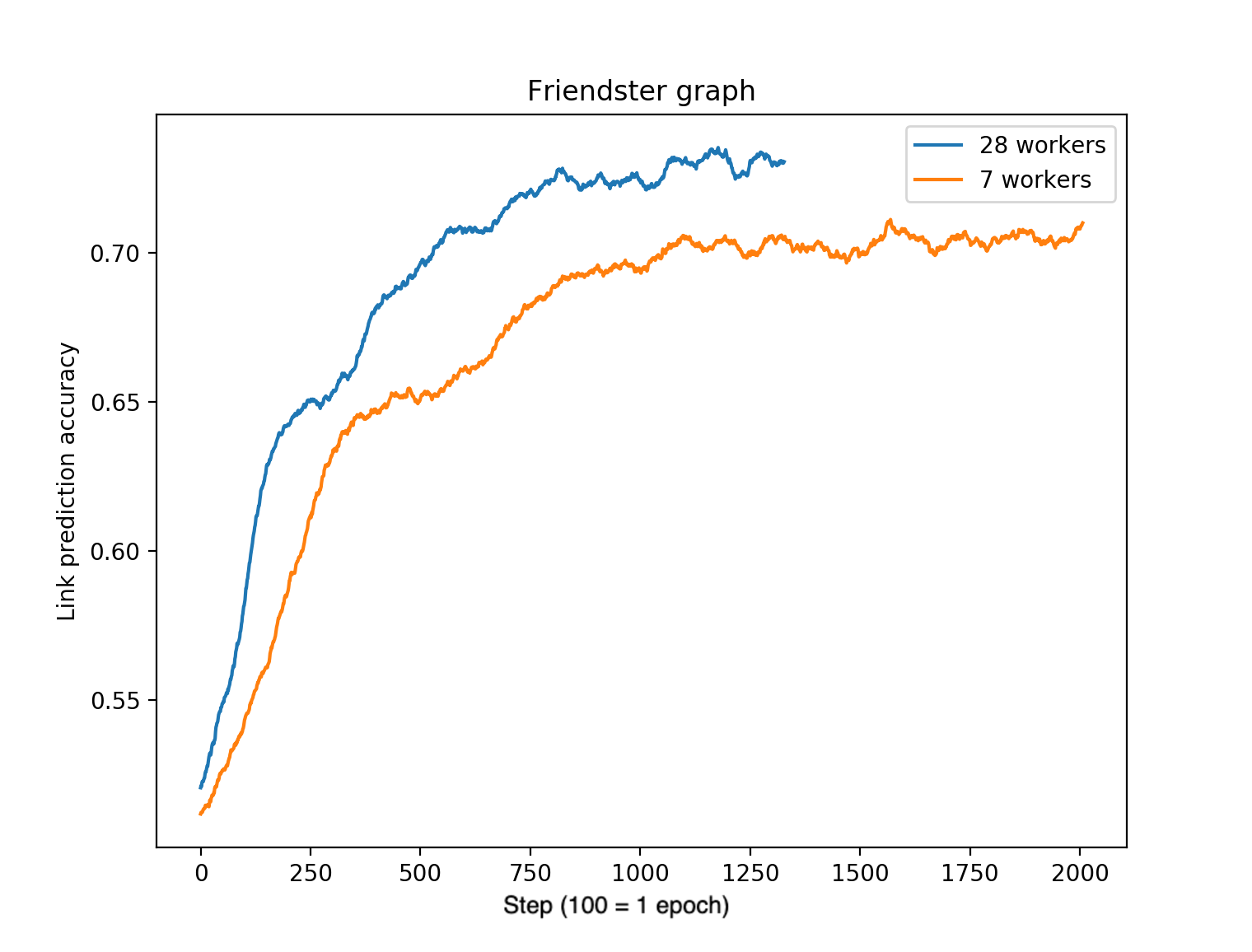}
        \caption{Extending training past 10 epochs demonstrates the benefits from using multiple workers in our system's distributed setup.}
        \label{fig:large_worker_v_small_worker}
    \end{subfigure}

\caption{Experiments run on the Friendster graph} 
\end{figure}

Table \ref{table:accuracies} presents a positive correlation between running training jobs on the Friendster dataset with more workers and link prediction accuracy. For example, training on the Friendster graph with 28 worker nodes results in the highest 72\% link prediction accuracy out of all experiments. Even more convincing is seeing that the proposed system - regardless of worker quantity - is able to predict the majority of positive and negative pairs in the testing set with similar accuracy. This balance demonstrates that our system is able to scale a commonly used embedding algorithm (skip-gram in these experiments) to find unique vector representations for users that are separated by some distance in embedding space, even in a dataset of nearly 65 million nodes. On the smaller IHG dataset, our system achieves 90\% total accuracy while training on the IHG graph, scoring a healthy balance of 95\% and 85\% positive and negative accuracy respectively. We attribute our system's better performance on this dataset to training with multiple vertex types, which essentially amount to additional features that skip-gram can use for deriving relationships between vertices.

Looking at Figure \ref{fig:many_workers_vs_link_fig} we also see that training with more workers also speeds up the rate of convergence on the Friendster graph. For example, a training job with 28 workers reaches 70\% link prediction accuracy by roughly 500 steps, while training with either 21 or 14 workers only crosses this threshold until the $550^{\text{th}}$ and $710^{\text{th}}$ step respectively. A system that only uses 7 workers is unable cross this threshold until well after the 1000th step, as shown in Figure \ref{fig:large_worker_v_small_worker}. These results demonstrate a few key insights about our proposed architecture. 

Firstly, a system with more workers is providing more gradient updates to the centralized parameter server at every training iteration. Receiving gradient signal from more neighborhoods within the graph is equivalent to updating the embedding vectors for all nodes with more topographic information about the graph that is outside their immediate vicinity. We see this effect in Figure \ref{fig:many_workers_vs_link_fig} at roughly step 300 (epoch 3), where all workers converge to roughly 65\% link prediction accuracy before fanning out into different final values. At the beginning of training, the system likely focuses on updating vertex embeddings with local topographic information. Training for longer, however, ties these localized updates with signal from other areas of the graph and explains the bump in accuracy scores after step 300 of training. Furthermore, horizontally scaling our system with more machines leads to more globally regularized (regularized due to asynchronous training) updates to all embedding vectors at every iteration. Thus, more workers accelerates the tying of local and global topographic information, and further explains the separation at step 300 in Figure \ref{fig:many_workers_vs_link_fig}. Figure \ref{fig:large_worker_v_small_worker} reinforces this argument by showing how a system with a large number of workers is able to train more swiftly and accurately over an extended training cycle when compared to one with only a few worker nodes.

Secondly, these results demonstrate the effectiveness of using our system to scale noise-contrastive estimation for computing embedding representations for millions of different vertices. We see in Figure \ref{fig:many_workers_vs_link_fig} that all training jobs, regardless of the number of available workers, are able to steadily increase in link prediction accuracy and eventually converge to an acceptable solution. Figure \ref{fig:large_worker_v_small_worker} shows that combining the distributed benefits of using more worker nodes with a NCE loss objective leads to a superior solution. Furthermore, our system's strong performance on the IHG dataset demonstrates an ability to successfully scale the existing skip-gram algorithm on graphs with several vertex types.

%% file: section-conclusion.tex
\section{Conclusion}
\label{sec:conclusion}

In this paper, we present a distributed architecture for training embeddings on large-scale graphs. We first explored the infrastructural advantages and disadvantages of other systems in industry that compute recommendations or embeddings at scale. We then presented our own approach, and delved into experimental results from scaling the skip-gram algorithm with our system on the Friendster and IHG graphs.

Our proposed system involves a few key features. Firstly, by transforming an input graph into a matrix of fixed-size random walks, we are able to avoid performing costly graph partitioning techniques and are instead able to efficiently row-wise partition a data matrix across many different worker nodes. Secondly, our introduction of parameter servers by vertex entity for storing embedding vectors allows each worker node to only fetch the required embeddings for a given set of vertices in a batch of data. Thus, our system does not require every worker to store a replica of the global TensorFlow computational graph and uses the highly efficient indexing operations described in Section 3.2 for retrieving and updating only subsets of all embeddings at every training iteration. This allows the model to use GPUs for training the embeddings by limiting the size of the computational graph to fit in GPU memory. At the same time it takes advantage of CPU and RAM for maintaining and serving the central embedding matrix.

We quantitatively show that scaling the number of worker nodes in our system leads to a positive correlation with link prediction accuracy on the Friendster dataset, and a 90\% total accuracy on the IHG graph. Our results also demonstrate that a larger number of workers not only speeds up the rate of training convergence but also maintains the quality of our resulting embedding vectors. 

There are several future directions for this work. We hope to abstract our system's training pipeline to work with more than two entities within a large-scale heterogeneous dataset. This will allow us to empirically evaluate our infrastructure against other heterogeneous systems in literature - e.g metapath2vec, HetNet - and compare performance with different combinations of vertex types. We also want to benchmark our system's performance on a variety of other quantitative metrics, including measuring average network latency across worker nodes, and measuring our system's overall training speed in comparison to other open-source industrial systems. Lastly, we hope to compile visualizations of our resulting embeddings to demonstrate our system's ability to maintain quality even while training at scale.